\documentclass[letterpaper, 10 pt, conference]{ieeeconf}  

\IEEEoverridecommandlockouts                              

\usepackage{graphics} 
\usepackage{epsfig} 
\usepackage{mathptmx} 
\usepackage{times} 
\usepackage{amsmath} 
\usepackage{amssymb}  
\usepackage{algorithm}
\usepackage[noend]{algpseudocode}
\usepackage{lipsum}
\usepackage{mwe}
\usepackage[noadjust]{cite}
\usepackage{filecontents}

\title{\LARGE \bf
Recursive Regression with Neural Networks: Approximating Hamilton-Jacobi-Isaacs PDE Solutions$^{*}$}

\author{Vicen\c{c} R\'{u}bies Royo$^{\dagger}$ and Claire Tomlin
\thanks{* This research is supported by ONR under the Embedded Humans MURI (N 00014-16-1-2206)}
\thanks{$^{\dagger}$
The authors are with the Department of Electrical Engineering and Computer Sciences, University of
California, Berkeley. \{vrubies, tomlin\}@berkeley.edu
}
}

\begin{document}

\maketitle
\thispagestyle{empty}
\pagestyle{empty}

\begin{abstract}
\nocite{*}
The majority of methods used to compute approximations to the Hamilton-Jacobi-Isaacs partial differential equation (HJI PDE) rely on the discretization of the state space to perform dynamic programming updates. This type of approach is known to suffer from the \textit{curse of dimensionality} due to the exponential growth in grid points with the state dimension. In this work we present an approximate dynamic programming algorithm that computes an approximation of the solution of the HJI PDE by alternating between solving a regression problem and solving a minimax problem using a feedforward neural network as the function approximator. We find that this method requires less memory to run and to store the approximation than traditional gridding methods, and we test it on a few systems of two, three and six dimensions.

\end{abstract}

\section{INTRODUCTION}
\label{sec:one}

Artificial neural networks are remarkable function approximators used in a myriad of applications ranging from complex controllers for robotic actuation \cite{Levine2016}, \cite{Schulman2015} to simple image classifiers for digit recognition \cite{LeCun1989} . They even find uses in physics to find approximations to solutions of PDEs and systems of coupled ordinary differential equations (ODEs) \cite{Lagaris1998}. Their success is in part achieved by their property of being universal function approximators \cite{Hornik1989}. 

In order to train a neural network one usually defines a cost function which captures the ``goodness" of the choice of parameters in the model, and uses gradient descent/ascent algorithms to improve them. In supervised learning, for example, cost functions such as the mean squared error or the mean absolute error are used to measure the discrepancy between input-output data pairs; unfortunately, in many cases one does not have access to such pairs, which limits the applicability of this approach. For instance, in many approximate dynamic programming settings, one would like to find an optimal control policy which minimizes some cost criterion of an agent operating in an environment. This function is usually unknown \textit{a priori}, so this problem can't be directly framed as a regression problem using input-output pairs. However, as seen in \cite{Mnih2013}, a neural network can be trained to approximate a Q-function\footnote{a Q-function is a type of value function used in reinforcement learning which expresses the value of taking a certain action at a given state given a reward structure \cite{Watkins1992} }  by generating a series of loss functions of the form

\begin{equation}
\label{eq:Q_learn}
L_i(\theta_i) = \mathbb{E}_{s,a \sim \rho} [(y_i - Q(s,a;\theta_i))^2].
\end{equation}
Here, the targets $y_i$ are generated from the same network that is being used to approximate the Q-function, hence the neural network has two purposes: approximation and target generation. In this work, we show that this idea can be extended to the domain of approximating solutions to HJI PDEs. More concretely, we present an algorithm which uses a feedforward neural network to approximate the desired solution by alternating between performing regression updates and generating new targets for regression by solving an optimal control problem. We then validate our formulation empirically by testing it on domains of two, three and six dimensions, and compare our results to current state-of-the-art numerical tools. 

The remainder of the paper is structured as follows: 
\begin{itemize}
\item in section \ref{sec:two}, we give a brief introduction to the HJI PDE and reachability theory.
\item in section \ref{sec:three}, we present some of the literature on PDE numerical methods using neural networks.
\item in section \ref{sec:four}, we present our algorithm for computing an approximation to the HJI PDE.
\item in section \ref{sec:five}, we show several experiments in domains of various sizes.
\item finally, we provide a discussion of the results in section \ref{sec:six}.
\end{itemize}

\section{Background}
\label{sec:two}
\subsection{The Hamilton-Jacobi-Isaacs PDE}
Let $V: \mathbb{R}^n \times [T,0] \to \mathbb{R} $ be a map, where $T \leq 0$. Then, given a time invariant system of the form $\frac{dx}{dt} = f(x,a,b)$ and boundary condition $V(x,0) = l(x)$, where $x \in \mathbb{R}^n$ is the state vector and $a \in A \subseteq \mathbb{R}^{m_a}$ and $b \in B \subseteq \mathbb{R}^{m_b}$ are inputs to the system\footnote{$a$ is usually taken to be the user input and $b$ is taken to be some bounded input disturbance}, we wish to find the solution to the minimum-payoff HJI PDE, associated to the reachability problem:

\begin{equation}
\frac{\partial V(x,t)}{\partial t} = - min \{0 , H(x,\nabla_x V) \},
 \label{eq:HJIPDE}
\end{equation}
where

\begin{equation}
H(x,\nabla_x V) := \underset{a \in A}{max} ~ \underset{b \in B}{min} ~ \nabla_x V^T f(x,a,b)
 \label{eq:Hamiltonian}
\end{equation}
is the Hamiltonian. The boundary condition $V(x,0) = l(x)$ encodes in its zero sub-level set (i.e. $l(x) \leq 0$) the region of interest in our state space known as the target set $\mathcal{T}$, and the solution $V(x,t)$ to (\ref{eq:HJIPDE}) encodes all the starting states whose induced trajectories starting at time $t$ will enter (and possibly leave) $\mathcal{T}$ within $|t|$, given the dynamics and input signals . 

More precisely, for some starting state $x_0$ and $t \leq 0$, $V(x_0,t) < 0$ if and only if the trajectory starting from $x_0$ at time $t$ enters $\mathcal{T}$ within $|t|$. 

\subsection{Safety and Optimal Control}
For every $t \in \mathbb{R}^{-} $, $V(x,t)$ contains in its zero sub-level set the set of states from which the disturbance can drive the system into $\mathcal{T}$ despite the best possible control. Thus, viewing the disturbance as a player who optimally tries to drive the system into the target set, and given some bounds on this disturbance, we can derive safety guarantees. This type of safety formulation is particularly useful for safety-critical systems \cite{Lygeros1999},\cite{Chen2015},\cite{Chen2016},\cite{Gillula2010}, where the target set is taken to be a dangerous region in our state space where a collision or a dangerous mode of operation might occur.

Last but not least, it is important to mention that the solution $V(x,t)$ also encodes the optimal inputs for the control and the disturbance by computing the arguments that maximize and minimize the Hamiltonian respectively. That is, given our solution $V(x,t)$, the optimal inputs for the control and the disturbance at time $t$ are given by
\begin{equation}
(a^*,b^*) = \underset{a \in A}{argmax} ~ \underset{b \in B}{argmin} ~ \nabla_x V^T f(x,a,b).
\label{eq:argmaxmin}
\end{equation}

\section{Neural Approximation of PDE Solutions}
\label{sec:three}

The problem presented in section \ref{sec:two} is in general not straightforward to solve and in many cases classical solutions do not exist. For this reason, trying to find an approximation instead of the actual solution can be a reasonable approach. The majority of numerical tools used to approximate solutions of PDEs, including (\ref{eq:HJIPDE}), use gridding techniques whereby finite differences are used to iteratively update values of $V(x,t)$ on a grid \cite{Mitchell2005},\cite{Mitchell2007}. Another approach to approximate solutions \cite{Lagaris1998} of general PDEs consists in training a feedforward neural network to solve the boundary value problem

\begin{equation}
\begin{split}
& G(x, \psi(x), \nabla \psi(x), \nabla^2 \psi(x)) = 0\\
& \psi(\tilde{x}) = A(\tilde{x}),
\end{split}
\label{eq:loss_gen_pde}
\end{equation}
by minimizing the loss function

\begin{equation}
L_\theta := \sum_{i=1}^N ~ G(x_i, \psi_\theta(x_i), \nabla \psi_\theta(x_i), \nabla^2 \psi_\theta(x_i))^2.
\label{eq:gen_pde}
\end{equation}
The subscript $\theta$ denotes the set of parameters that define the candidate solution $\psi_\theta(x)$; if the candidate solution uses a neural network approximator, as we will see in subsection \ref{sub:two}, $\theta$ represents a set of matrices and vectors.  Additionally, the candidate solution is defined as $\psi_\theta(x) := A(x) + F(x,N_\theta(x))$, a form which by construction satisfies the boundary condition. This is accomplished by enforcing that $F(x,N_\theta(x)) = 0$ as $x \to \tilde{x}$. Finally, $N_\theta(x)$ is a feedforward neural network, which we will succinctly define in section \ref{sec:four}.
 
Although this approach might be well suited for some problems, it still requires the discretization of the domain. Moreover, special care must be taken when computing the gradient of the loss with respect to the parameters. For instance, following the previous procedure, the loss function for the HJI PDE would be written as 

\begin{equation}
\label{eq:HJI_error}
L_\theta := \sum_{i=1}^N (\frac{\partial V(x_i,t_i)}{\partial t} + min \{0 , H(x_i,\nabla_x V) \})^2,
\end{equation}
but the $min$ inside makes this expression not differentiable everywhere. Using the above expression for the loss function, Djeridane et al. \cite{Djeridane2006} show that in cases where $H(x_i,\nabla_x V)$ can be computed explicitly one can directly define an expression for the gradient. In their work, they also show that the points used to compute the loss need not be taken from a grid, and instead choose to acquire them via random sampling. However, trying to approximate the solution of (\ref{eq:HJIPDE}) using this type of loss function doesn't work well in practice (see the example in the Appendix).

In this work, we tackle the problem of finding an approximate solution to (\ref{eq:HJIPDE}) from a different perspective. Whereas (\ref{eq:HJI_error}) is designed to satisfy the PDE by penalizing deviations of the residual $G$ away zero, we turn the problem into a recursive regression problem with loss functions similar to (\ref{eq:Q_learn}). In the next section we explain the algorithm in detail.

\section{Recursive Regression Algorithm}
\label{sec:four}

In this section we will present the algorithm which is at the core of this work. We use the term recursive regression to emphasize the fact that we are not solving a typical regression problem with fixed input-output pairs, and that instead we are repeatedly generating new regression targets as the algorithm proceeds. Furthermore, we consider the algorithm to fall in the realm of approximate dynamic programming algorithms, since we are using a parametric function to approximate the solution $V(x,t)$ \cite{Powell2009},\cite{Powell2011}. In the following subsections we bring forth some important definitions that will help introduce the algorithm at the end. 

\subsection{Preliminaries}
\label{sub:one}

Equation (\ref{eq:HJIPDE}) can be used to derive an approximate expression for $V(x,t)$ by rewriting the partial derivative with respect to time as a finite difference:

\begin{equation}
\frac{V(x,t)-V(x,t-\Delta t)}{\Delta t} \approx - min \{ 0, \underset{a}{max} ~ \underset{b}{min} ~ \nabla_x V^T f(x,a,b) \},
\end{equation}
where $\Delta t$ is assumed to be small. Rearranging terms we see that

\begin{equation}
V(x,t-\Delta t) \approx min \{ V(x,t) , \underset{a}{max} ~ \underset{b}{min} ~ V(x + f(x,a,b)\Delta t, t) \}.
\end{equation}
In the next subsections we show how this approximation becomes relevant.

\subsection{Candidate Approximation}
\label{sub:two}

Let $N_\theta(x,t):\mathbb{R}^n \times \mathbb{R} \to \mathbb{R}$ be a feedforward neural network with $k$ hidden layers and identity activation for the output layer, then 

\begin{equation}
N_\theta(x,t) := l_{k+1} \circ l_{k} \circ ... \circ l_{1}(x,t)
\end{equation}
where $\circ$ is the composition operator and $l_{j}(v) := \sigma(A_jv + b_j)$. Here, $v$ is the output of $l_{j-1}$ (the previous layer), and $A_j \in \mathbb{R}^{m_j \times dim(v)}$ with $b_j \in \mathbb{R}^{m_j}$ define an affine transformation on $v$. In addition, $\sigma: \mathbb{R}^{m_j} \to \mathbb{R}^{m_j}$ is the activation function, a fixed non-linear function which operates component-wise on its input. For instance, the sigmoid function $\sigma(x) = 1/(1 + e^{-x})$ or the softplus function $\sigma(x) = ln(1+e^x)$ are examples of commonly used activations. The last activation function for $l_{k+1}$ is the identity. Finally, we define $\theta := \{A_{k+1},...,A_{1},b_{k+1},...,b_{1}\}$ to be the set of parameters of the neural network.

Now that we have defined the type of approximator we will use, we define, similar to \cite{Lagaris1998}, the form of our candidate approximation to be

\begin{equation}
V_\theta(x,t) := V(x,0) + t N_\theta(x,t),
\label{eq:candid}
\end{equation}
where as $t \to 0$, $V_\theta(x,t) = V(x,0)$, thus satisfying the boundary conditions by construction.

\subsection{Loss functions}
\label{sub:three}
With the information from the previous two subsections, we can now define a series of loss functions similar to (\ref{eq:Q_learn})

\begin{equation}
L_{i}(\theta_{i}) := \frac{1}{N} \sum_{j=1}^N | y_j - V_{\theta_{i}}(x_j,t_j-\Delta t) |
 \label{eq:loss_exp}
\end{equation}
for $i = 1,2,...~$, where $N$ corresponds to the number of points sampled from some probability distribution defined over $S \times [T,0]$, where $S$ is some compact set in $\mathbb{R}^n$ containing the target set $\mathcal{T}$. For the targets,

\begin{equation}
y_j = min \{ V_{\theta_{i-1}}(x_j,t_j) , \underset{a}{max} ~ \underset{b}{min} ~ V_{\theta_{i-1}}(x_j + f(x_j,a,b)\Delta t, t) \}.
\end{equation}
Here, $\theta_0$ is the initial set of parameters, which is taken over some normal distribution with mean zero. The benefit of defining these series of loss functions is the following: it allows us to break the problem into two intuitive subtasks, where the first subtask consists in sampling $N$ points, solving an optimization problem and generating targets $y_j$, and the second subtask consists in performing regression over the targets $y_j$ by using gradient descent. Once several regression updates have been taken, we fix the model and repeat. What we mean here by ``fixing" the model, is that once the regression subtask is complete, yielding a new $\theta_i$, we immediately increase the value of $i$ by one, thus making $\theta_i$ and older model from which we will sample $N$ points to train $\theta_{i+1}$. In reality, as we will see in the algorithm, there is no need to explicitly define the iterator $i$, but it helps understand the order of events.

\subsection{Algorithm}
\label{sub:four}

\begin{algorithm}[]
  \caption{Recursive Regression}
  \begin{algorithmic}[1]
  	\State \textbf{Input:} $V(x,0)$, $f(x,a,b)$, A, B, $S$, T, $N$, interval, $K$(batch size), $\gamma$ (momentum decay), $\eta$ (learning rate), $stop$
  	\State Set $\Delta t$ small
  	\State $iter \gets 0$, $\nu \gets 0$
 	\State $\theta \sim \mathcal{N}(0,0.1I)$
 	\State Define $V_\theta(x,t) := V(x,0) + t N_\theta(x,t)$
 	\State Define $L_\theta := \frac{1}{K}\sum_{k=0}^K |y_k - V_\theta(x_k,t_k-\Delta t)|$ 
 	\While{iter $< stop$}
 		\If{mod(iter,interval) == 0}
 			\State $R \gets ~ $ empty array of size $N$
 			\State Sample $N$ pairs $(x,t) \sim Uniform(S \times [-T,0])$
 			\For{$j=0$ to $N$}
 				\State $(a_j^*,b_j^*) \gets \underset{a  \in A}{argmax} ~ \underset{b \in B}{argmin} ~ \nabla_x \hat V_{\theta}^T f(x,a,b)$
 				\State $\tilde{x}_j \gets$ Simulate($x_j,a^*,b^*,\Delta t$)
 				\State $y_j \gets min\{V_\theta(x_j,t_j), V_\theta(\tilde{x}_j,t_j)\}$
 				\State $R_j \gets ((x_j,t_j),y_j)$
 			\EndFor
 		\EndIf
 	\State $b \gets K$ elements from $R$ picked at random
 	\State $\nu \gets \gamma \nu + \eta \nabla_\theta L_\theta(b) $
 	\State $\theta \gets \theta - \nu$
 	\State $iter \gets iter+1$
 	\EndWhile 
 	\State \textbf{Output:} $\hat V_\theta(x,t)$

	\end{algorithmic}
	\label{algo:dpnn}
\end{algorithm}
   
In this section we present the algorithm for Recursive Regression in pseudo-code format. In Algorithm \ref{algo:dpnn}, the inputs correspond, in order, to the boundary condition $V(x,0)$, the dynamics $f(x,a,b)$, the set of control inputs A and disturbance inputs B, the set of states $S \supset \mathcal{T}$ and the time horizon $T \leq 0$ over which we compute the approximation. $N$ corresponds to the number of points we sample over $S \times [T,0]$ to generate the targets and the variable 'interval' encodes the number of regression steps/updates we perform on the parameter vector $\theta$ before resampling. The batch size $K$ and momentum decay $\gamma$ hyper-parameters are used to implement stochastic gradient descent with momentum, a special type of gradient descent algorithm which usually converges faster than batch gradient descent \cite{Qian1999},\cite{Bottou2010}. Finally, the input $stop$ determines the number of iterations we run before halting the algorithm.

The first few steps of the algorithm are used to initialize the parameters of the neural network and to initialize the approximation $V_\theta$ and the loss function $L_\theta$. Once this is finished, we loop through the while-loop for number $stop$ iterations. As one can see, we only enter in the if-statement when the iterator $iter$ is a multiple of the input $interval$; at that point we renew the regression points by emptying the array $R$ and sampling $N$ new $(x,t)$ pairs uniformly over $S \times [T,0]$. Using our model $V_\theta(x,t)$, we then iterate through all the samples, compute the optimal inputs to take, generate the targets for regression and fill the array $R$ with those new values. Finally, outside of the if-statement, we run the regression updates by sampling $K$ elements at random from $R$ and performing a gradient descent step. We repeat these regression updates until $iter$ is again a multiple of $interval$.

Algorithm \ref{algo:dpnn} can be viewed as a bootstrapping method in that lines 12,13 and 14 make use of $V_\theta(x,t)$ to generate new points for regression to train $N_\theta(x,t)$ which in turn modifies $V_\theta(x,t)$ itself. At first glance, it is unclear whether the generated pairs $((x_j,t_j),y_j)$ will result in a good approximation to the solution of our PDE by repeatedly doing regression; however, given the form of our candidate function (\ref{eq:candid}) we expect that points sampled near $t = 0$ will in fact be reasonable approximations of $V(x,t)$ for small $t$. Given this assumption, we hypothesize that despite the presence of misleading data, our network will still be able to do a good job at regressing over all points, thus improving our initial model and allowing the generation of improved data. By repeating this procedure, we expect that values from the boundary will ``propagate" backward in time (possibly with some minor error) as the algorithm proceeds.

Another important aspect from line 13 is the assumption that we can simulate the dynamics forward in time. For the experiments in the next sections a Runge-Kutta method with 4 stages (RK4) was used, although other methods can be used as well. 

\section{Numerical Results}
\label{sec:five}

In this section we present 2, 3 and 6-dimensional experiments to empirically assess the validity of our formulation. To measure the performance of the algorithm, we compare the output of our approximation at various learning stages against accurate approximations taken from state-of-the-art tools. In particular, we make use of the LevelSet Toolbox \cite{Mitchell2007}, a powerful computational tool for obtaining approximations to Hamilton-Jacobi (HJ) PDEs using gridding methods. All experiments were performed on a machine with 12 Intel(R) Core(TM) i7-4930K, 3.40GHz processors.

\subsection{Error Metrics}

The first error metric to be used will be

\begin{equation}
\label{eq:errorm1}
E_1(V_\theta(x,t)) := \frac{1}{M} \sum_{i=1}^M |V(x_i,t_i) - V_\theta(x_i,t_i)|
\end{equation}
where $M$ is the number of points chosen from our domain to compute the average absolute error. Unfortunately, since the solution $V(x,t)$ is typically unknown, we have to resort to an accurate approximation instead. Finally, we also use the PDE residual as a second error metric 

\begin{equation}
\label{eq:errorm2}
E_2(V_\theta(x,t)) :=   \frac{1}{M} \sum_{i=1}^M | \frac{\partial V_\theta(x_i,t_i)}{\partial t} + min \{0 , H(x_i,\nabla_x V_\theta) \} |
\end{equation}
similar to the one defined in (\ref{eq:HJI_error}), which denotes the extent by which (on average) the approximation is violating the PDE equality. Here the residual error is taken solely as a metric to assess the progression in learning and not as a metric to assess the accuracy of the approximation. For all experiments $M=3000$, with all points chosen uniformly at random over $S \times [T,0]$. 

\subsection{Pursuit-Evasion Game in 2D}
\label{exp:one}

In this experiment we explore a pursuit-evasion game in 2 dimensions where a pursuer has to intercept an evader. In a simplified approach, we assume the evader is constrained to move in one dimension, whereas the pursuer has the liberty to move in two dimensions by changing the angle of its heading. Fixing the evader at the origin with its heading aligned with the x-axis we frame the problem in relative coordinates between the evader and pursuer, that is $x = [x_r ~ y_r]^T$, where $x_r$ and $y_r$ represent the $x$ and $y$ position of the pursuer relative to the evader's frame of reference. This system's dynamics are readily encoded in the following equation

\begin{equation}
\label{eq:pursuer_evader}
\begin{bmatrix}
\dot{x}_r\\
\dot{y}_r
\end{bmatrix} = f(x,a,b) = \begin{bmatrix}
v_p cos(b) - a\\
v_p sin(b)
\end{bmatrix}
\end{equation}
Here we encode the capture condition by defining $V(x,0) = ||x||_2 - 1$ (when the evader and pursuer are one unit of distance apart) and we define the time horizon to be $T=-1.0$. We also assume $v_p = 2.0$, $a \in [-2,2]$ and $b \in [0,2\pi]$. For this experiment, a feedforward neural network with a single hidden layer of 10 units and sigmoid activation functions was used. The number of points sampled was chosen to be $N = 500$, uniformly picked over the set $S := \{(x_r,y_r) | x_r,y_r \in [-5,5] \}$ and over $t \in [T,0]$. The batches were picked to be of size $K = 10$, momentum decay $\gamma = 0.95$ and learning rate $\eta = 0.1$. The interval to renew the regression points was chosen to be $1000$ iterations and the program was halted at 300,000 iterations.

\begin{figure}[t]
\begin{center}
\centering
    \includegraphics[height=1.0\linewidth, width=1.0\linewidth]{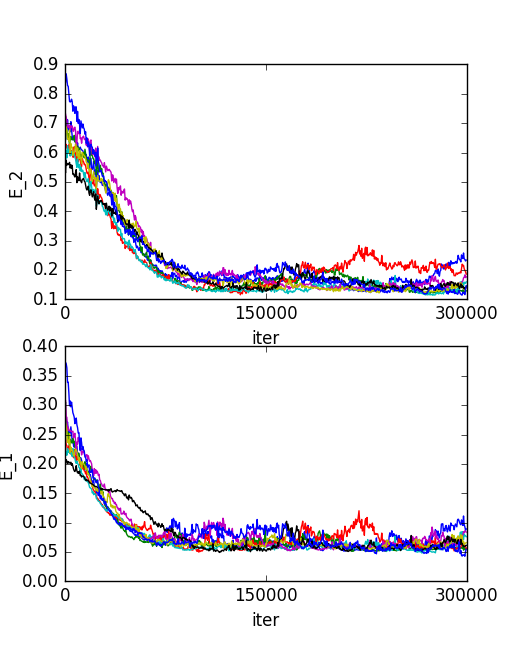}
\end{center}
\caption{The top figure shows the mean absolute PDE error $E_2$ and the second shows the mean absolute error $E_1$ for the 2D pursuit-evasion game for each of the 8 threads. For $E_1$, one can see that as the number of iterations increases both error metrics decrease for all threads until they reach some a value near $0.06$.}
\label{fig:lin_pevader_plot}
\end{figure}
\begin{figure}[t]
\begin{minipage}[t]{0.47\linewidth}
    \includegraphics[width=\linewidth]{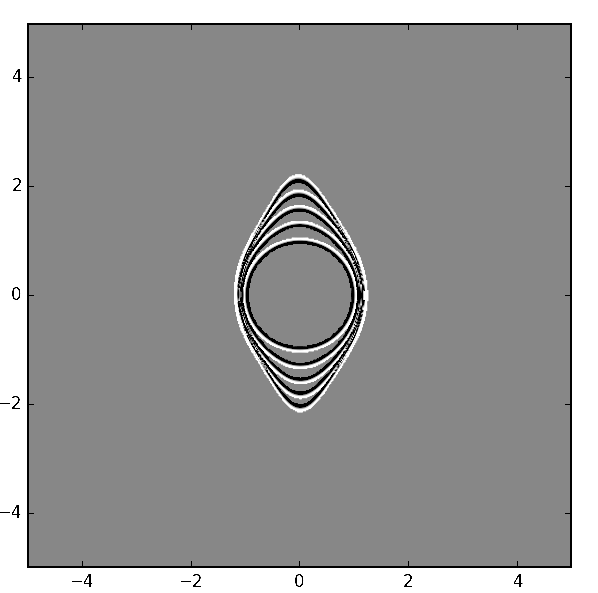}
    \caption{The zero level sets at $t=0,-0.25,-0.5, -0.75 $ and $ -1.0$ computed using the neural network approximation.}
    \label{fig:nn2d}
\end{minipage}%
    \hfill%
\begin{minipage}[t]{0.47\linewidth}
    \includegraphics[width=\linewidth]{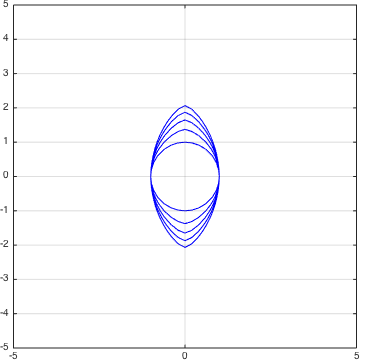}
    \caption{The zero level sets at $t=0,-0.25,-0.5, -0.75 $ and $ -1.0$ computed by the LevelSet Toolbox using a grid of 51 points per axis.}
    \label{fig:grd2d}
\end{minipage} 
\end{figure}

The results shown in Fig. \ref{fig:lin_pevader_plot} where taken over 8 concurrent runs of the algorithm. The overall time to run the 300,000 iterations over all threads was $443$ seconds. The average $E_1$ error at halting time was in the order of $6 \times 10^{-2}$, whereas the $E_2$ error was in the order of $1.5 \times 10^{-1}$. The points used to compute $E_1$ were taken from a $51 \times 51$ grid at $t=-0.5$ of a pre-computed approximation using the LevelSet Toolbox. Finally, we also include a visual comparison in Fig. \ref{fig:nn2d} and \ref{fig:grd2d} between the zero level sets computed by the neural network and the ones computed with the LevelSet Toolbox at $t=0,-0.25,-0.5,-0.75$ and $-1.0$.


\subsection{Pursuit-Evasion Game in 3D}
\label{exp:two}

This experiment is similar to the 2 dimensional example, but with the twist that now the evader can also control its heading. This time, however, neither the evader nor the pursuer can control the angle directly, and instead can only control the rate of rotation, thus making the problem 3 dimensional. Again, we frame the problem in relative coordinates from the evader's frame of reference. This dynamical system can be written as follows:

\begin{equation}
\label{eq:pursuer_evader}
\begin{bmatrix}
\dot{x}_r\\
\dot{y}_r\\
\dot{\theta}_r
\end{bmatrix} = f(x,a,b) = \begin{bmatrix}
-v_e + v_p cos(\theta_r) + ay_r\\
v_p sin(\theta_r) - a x_r\\
b - a\\
\end{bmatrix}
\end{equation}

For this problem the capture condition is encoded in the boundary condition $V(x,0) = ||[x_r ~ y_r]^T||_2 - 1$ (where we ignore $\theta_r$ since the capture condition only depends on the distance) and we consider a the time horizon $T = -1.0$. In this problem we give both pursuer and evader the same speed $v_p = v_e = 1.0$ and the same turning rates $a,b \in [-1,1]$. Unlike the previous experiment, we used a neural network with two hidden layers of 10 and 5 units respectively and sigmoid activations. The number of points sampled was chosen to be $N = 2000$, uniformly picked over the set $S := \{(x_r,y_r,\theta_r) | x_r,y_r \in [-5,5], \theta_r \in [-\pi,\pi] \}$ and over $t \in [T,0]$. The batches were picked to be of size $K = 25$, momentum decay $\gamma = 0.999$ and learning rate $\eta = 0.001$. The interval to renew the regression points was chosen to be $1000$ iterations and the program was halted at 1,000,000 iterations. 

\begin{figure}[t]
\begin{center}
\centering
    \includegraphics[height=1.0\linewidth, width=1.0\linewidth]{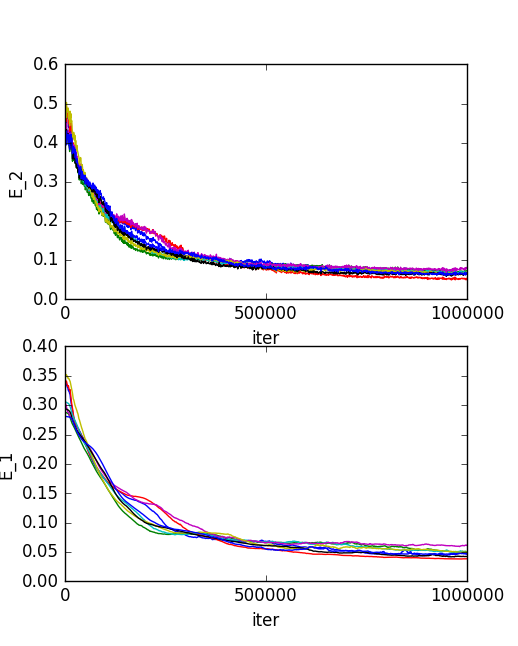}
\end{center}
\caption{The top figure shows the mean absolute PDE error $E_2$ and the second shows the mean absolute error $E_1$ for the 3D pursuit-evasion game for each of the 8 threads. Similar to the 2D case, one can see that as the number of iterations increases the error metric $E_1$ decreases for all threads until they all reach a value near $0.05$.}
\label{fig:3dplots}
\end{figure}
\begin{figure}[t]
\begin{center}
\centering
    \includegraphics[height=1.0\linewidth, width=1.0\linewidth]{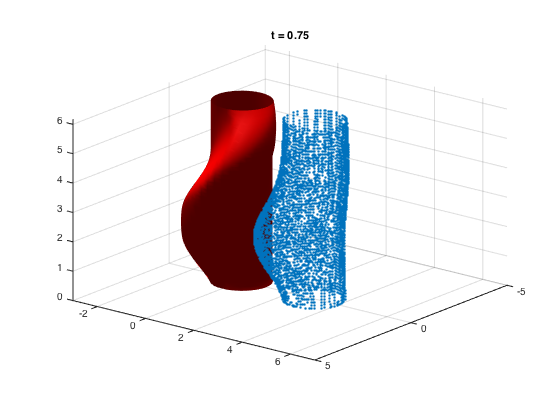}
\end{center}
\caption{Zero level set of $V(x,t)$ at $t=-0.75$ for the grid and the neural network. The surface on the left was obtained with the LevelSet Toolbox using a grid with 51 points in each dimension. The surface on the right (represented as point cloud) corresponds to points sampled near $V_\theta(x,-0.75) \approx 0$ using the neural network. The point cloud was shifted 3 units along the y-axis for display purposes.}
\label{fig:3dcomp}
\end{figure}

 
As shown in Fig. \ref{fig:3dplots}, both error metrics decrease as the algorithm progresses, reaching an average error for $E_1$ in the order of $5.0 \times 10^{-2}$ and an average error for $E_2$ in the order of $1.0 \times 10^{-1}$. The points used to compute $E_1$ were taken from a $51 \times 51 \times 50$ approximation grid at $t=-1.0$. This set of experiments was run concurrently using 8 threads and the total time for all threads to finish was $2448$ seconds. Finally, Fig. \ref{fig:3dcomp} shows a comparison of the zero level set at $t=-0.75$, which is now a 3D surface, between the gridding method and the neural network. 

\subsection{Pursuit-Evasion Game in 6D}
\label{exp:three}

Even though the  previous experiment is represented in relative coordinates, it seems more intuitive to define the position and heading of the pursuer and evader independently relative to a global frame of reference. Unfortunately, defining the states in this way leads to a problem with a 6 dimensional state space. As we know, gridding methods only scale up to 4 dimensions, and that is in part why the problem is usually expressed in 3 dimensions. However, as we will see in this last experiment, one can directly work with the 6 dimensional formulation and compute a good approximation using the algorithm.

An important reason for testing our algorithm with this specific high-dimensional dynamical system is that we already have the pre-computed approximation for the 3D experiment. The reason why this is useful is that given a state vector in $x \in \mathbb{R}^6$ for this system, one can transform it into relative coordinates and check that the high-dimensional approximation matches the lower dimensional one.

The dynamics for in this scenario are as follows:

\begin{equation}
\label{eq:pursuer_evader}
\begin{bmatrix}
\dot{x}_e \\
\dot{y}_e \\
\dot{x}_p \\
\dot{y}_p \\
\dot{\theta}_e \\
\dot{\theta}_p \\
\end{bmatrix} = f(x,a,b) = \begin{bmatrix}
v_e cos(\theta_e)\\
v_e sin(\theta_e)\\
v_p cos(\theta_p)\\
v_p sin(\theta_p)\\
a\\
b\\
\end{bmatrix}.
\end{equation}
It is important to point out that even though the dynamics have a nice structure, the algorithm doesn't alter the problem in any way. That is, if there exists some transformation of the inputs to ease the task of learning the approximation, it is up to the neural network to ``find" it via gradient descent

For this problem the target set is encoded in the boundary condition $V(x,0) = ||[x_p ~ y_p]^T - [x_e ~ y_e]^T||_2 - 1$ and we consider the time horizon $T = -1.0$. For this problem we give both pursuer and evader the same speed $v_p = v_e = 1.0$ and the same turning rates $a,b \in [-1,1]$. In this experiment we used a neural network with three hidden layers with 50 units each and softplus activations. The number of points sampled was chosen to be $N = 50000$, uniformly picked over the set $S := \{(x_e,y_e,x_p,y_p,\theta_e,\theta_p) | x_e,y_e,x_p,y_p \in [-15,15], \theta_e,\theta_p \in [0,2\pi] \}$ and over $t \in [T,0]$. The batches were picked to be of size $K = 100$, momentum decay $\gamma = 0.9999$ and learning rate $\eta = 0.0001$. The interval to renew the regression points was chosen to be $1000$ iterations and the program was halted at 1,000,000 iterations.  

\begin{figure}[t]
\begin{center}
\centering
    \includegraphics[height=1.0\linewidth, width=1.0\linewidth]{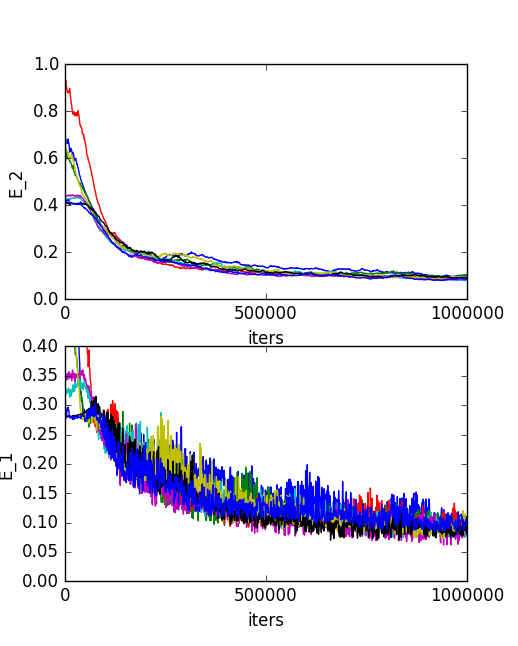}
\end{center}
\caption{The top figure shows the mean absolute PDE error $E_2$ and the second shows the mean absolute error $E_1$ for the pursuit-evasion game in 6D for each of the 8 threads. Similar to the previous two cases, one can see that as the number of iterations increases the error metric for $E_1$ decreases for all threads until they reach a value around $0.09$.}
\label{fig:6dplots}
\end{figure} 
As shown in Fig. \ref{fig:6dplots}, both error metrics decrease as the algorithm proceeds, reaching an average error for $E_1$ in the order of $9.0 \times 10^{-2}$ and an average error for $E_2$ in the order of $1.0 \times 10^{-1}$. To compute $E_1$, M points were sampled from the set $\{(x_e,y_e,\theta_e) | x_e,y_e \in [-5,5] , \theta_e \in [0,2\pi]\}$. Using the grid approximation from the 3D problem, the corresponding pursuer state $[x_p,y_p,\theta_p]$ was computed for each one of the M points. Finally, each pair of states was fed into the neural network and the output compared against the values in the 3D grid. This set of experiments was run concurrently using 8 threads and the total time for all threads to finish was 1 hour and 50 minutes.
\subsection{Self-Consistency}

Having the error $E_1$ decrease with the number of iterations indicates that points on the surface of the approximation are approaching points on the surface of the solution. However, since we have computed an approximation for the minimum-payoff HJI PDE, we can actually get a notion of ``self-consistency" of the approximation by comparing the minimum attained throughout a trajectory with respect to $V(x,0)$ and the payoff predicted by $V_\theta(x,t)$. That is, given $V_\theta(x,t)$, we can sample $N$ points in $S$, and starting at $t=-1.0$, simulate the trajectories forward in time while also keeping track of the minimum value attained with respect to $V(x,0)$. Once the minimum values have been computed, we can  then compare the discrepancy between these values and the value given by $V_\theta(x,t)$ at $t=-1.0$ by computing the average absolute difference over all $N$ points. Intuitively, this shows that the controller induced by $V_\theta$ generates trajectories which are consistent with the approximation itself. Unfortunately, due time constraints we were only able to compute the self-consistency for the first two systems. Tables \ref{tableone} and \ref{tabletwo} show the values for the self-consistency at the start and end of training with $N = 3000$. As one can see, once the model has been trained the self-consistency improves, indicating that the controller derived from $V_\theta(x,t)$ induces trajectories whose minimum payoff approaches the values given by directly querying the approximation.

\begin{table}[t]
\caption{Self-Consistency for 2D System}
\label{tableone}
\begin{center}
\begin{tabular}{|c|c|c|}
\hline
 & itr = 0 & itr = 300,000\\
\hline
Thread 1 & 0.518686 &  0.139768\\
\hline
Thread 2 & 0.371097 & 0.114652\\
\hline
Thread 3 & 0.447682 & 0.136796\\
\hline
Thread 4 & 0.353983 & 0.123748\\
\hline
Thread 5 & 0.462644 & 0.109586\\
\hline
Thread 6 & 0.400236 & 0.160209\\
\hline
Thread 7 & 0.604639 & 0.118796\\
\hline
Thread 8 & 0.508214 & 0.158434\\
\hline
\end{tabular}
\end{center}
\end{table}

\begin{table}[t]
\caption{Self-Consistency for 3D System}
\label{tabletwo}
\begin{center}
\begin{tabular}{|c|c|c|}
\hline
 & itr = 0 & itr = 1,000,000\\
\hline
Thread 1& 0.385512 & 0.109295\\
\hline
Thread 2 & 0.41547 & 0.110157\\
\hline
Thread 3 & 0.350484 & 0.11377\\
\hline
Thread 4 & 0.460214 & 0.126736\\
\hline
Thread 5 & 0.480284 & 0.116735\\
\hline
Thread 6 & 0.346113 & 0.107339\\
\hline
Thread 7 & 0.372726 & 0.129273\\
\hline
Thread 8 & 0.344426 & 0.110532\\
\hline
\end{tabular}
\end{center}
\end{table}

\section{DISCUSSION}
\label{sec:six}

We have seen in the previous section that the algorithm can successfully train a neural network to approximate the solution of the HJI PDE for problems of various sizes. In this section we discuss the strengths and limitations of this approach by comparing it to traditional gridding techniques.

\subsection{Memory Improvement}

Two of the main advantages of using this algorithm is the low amount of memory required to run it and to store the resulting approximation. For the experiments, most of the memory was employed to store the sampled points; since these points are used and eventually discarded, we only require a small amount of memory at runtime. Using gridding methods directly, the memory requirement grows exponentially with the state dimension.

In terms of storing the approximation, we can see that there are significant memory improvements using our approach. For instance, in experiment one, using a standard grid of $[51,51,10]$ discretization points per axis (i.e. $51$ in $x_r$, $51$ in $y_r$ and $10$ in t), $26,010$ values need to be stored. Using our approximation instead, we reduce the amount of storage needed to only $51$ weights. Similarly, for the second experiment we only need to store $111$ weights, as opposed to $51^3 \times 10$ values using a grid. Finally, in the last experiment we store the approximation in $5551$ weights rather than $51^6 \times 10$ for a regular sized grid.

\subsection{Runtime Issues}

One of the drawbacks of this approach for lower dimensional systems is the time it takes to converge to an approximation. Whereas gridding approaches typically take on the order of seconds or minutes to compute the approximation for 2 and 3 dimensional systems, training a neural network takes longer. However, since the time complexity for gridding algorithms is exponential with the state dimension, this only becomes an issue in lower dimensional settings.

\subsection{Parametric Approximation}

Another important aspect of the algorithm is the fact that it yields an approximation which is well-defined and differentiable everywhere. Whereas in gridding methods one needs to interpolate values that fall outside of the grid, using a neural network we can query points directly. Moreover, we can easily compute the derivative of the output with respect to any of the inputs exactly using the backpropagation algorithm.

\subsection{Future Work}

For now, one of the drawbacks of using this technique is the lack of theoretical guarantees. From the experiments we have done thus far, we have empirical evidence suggesting that the algorithm converges and can yield good approximations. However, to assess convergence and the accuracy of the approximation we resorted to using pre-computed solutions from the state-of-the-art, which is not ideal. Therefore, our main focus at the moment is to find these theoretical guarantees by looking into the statistical learning literature, especially because the end-goal is to use this algorithm for safety analysis, so having well-established bounds on the approximation error is crucial. 

Finally, some other research directions we are currently pursuing include the idea sampling smartly over the training domain rather than uniformly at random and investigating how the size of the training domain affects the resulting approximation. We believe that using normalization techniques our current results could be further improved. Similarly, since several models can be trained concurrently, we are interested in exploring whether random forests (the average output over all the trained models) improve the resulting approximation. Lastly, we are also investigating how the algorithm performs on GPUs and we hope to substantially improve the training time and be able to try larger and more complex neural architectures.




\section*{APPENDIX}

Here we show the learning curves using the exact same neural network as in experiment one, but employing the loss function defined in (\ref{eq:HJI_error}).
\begin{figure}[H]
\begin{center}
\centering
    \includegraphics[height=1.0\linewidth, width=1.0\linewidth]{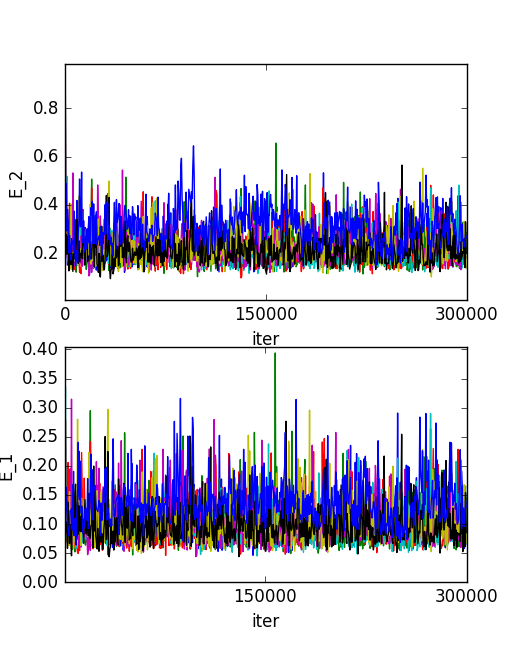}
\end{center}
\caption{The top figure shows the mean absolute PDE error $E_2$ and the second shows the mean absolute error $E_1$ for the 2D pursuit-evasion game using the alternate loss function for each of the 8 threads. For $E_1$, one can see that the error does not converge to any particular value and keeps oscillating.} 
\label{fig:last}
\end{figure}
As one can appreciate from Fig. \ref{fig:last}, the error seems to drop very sharply at the beginning and then oscillates erratically for all the threads, suggesting that the approximations are not converging.

\bibliographystyle{IEEEtran}
\bibliography{references}

\addtolength{\textheight}{-12cm}   

\end{document}